\documentclass[11pt,twocolumn,twoside]{article}
\usepackage{fully3d}
\usepackage{graphicx}
\usepackage{booktabs}
\usepackage{tabularx}
\usepackage{listings}
\usepackage{xcolor}

\usepackage{subcaption}

\lstdefinestyle{python}{
    language=Python,
    basicstyle=\ttfamily\small,
    keywordstyle=\bfseries\color{blue},
    stringstyle=\color{red},
    commentstyle=\itshape\color{gray},
    morekeywords={as}, 
    breaklines=true,
    frame=single,
    numbers=left,
    numberstyle=\tiny\color{gray},
    captionpos=b,
    showspaces=false,
    showstringspaces=false,
    tabsize=4
}

\begin{document}

\title{An update to PYRO‐NN: A Python Library for Differentiable CT Operators} 

\author[1]{Linda-Sophie~Schneider}
\author[1]{Yipeng~Sun}
\author[1]{Chengze~Ye}
\author[2]{Markus~Michen}
\author[1]{Andreas~Maier}

\affil[1]{Pattern Recognition Lab, Friedrich-Alexander University Erlangen-Nuremberg, Erlangen, Germany}

\affil[2]{Fraunhofer EZRT, Fürth, Germany}

\maketitle
\thispagestyle{fancy}


\begin{customabstract}
Deep learning has brought significant advancements to X-ray Computed Tomography (CT) reconstruction, offering solutions to challenges arising from modern imaging technologies. These developments benefit from methods that combine classical reconstruction techniques with data-driven approaches. Differentiable operators play a key role in this integration by enabling end-to-end optimization and the incorporation of physical modeling within neural networks.

In this work, we present an updated version of PYRO-NN, a Python-based library for differentiable CT reconstruction. The updated framework extends compatibility to PyTorch and introduces native CUDA kernel support for efficient projection and back-projection operations across parallel, fan, and cone-beam geometries. Additionally, it includes tools for simulating imaging artifacts, modeling arbitrary acquisition trajectories, and creating flexible, end-to-end trainable pipelines through a high-level Python API. Code is available at: \url{https://github.com/csyben/PYRO-NN}

%


\end{customabstract}

\section{Introduction}
X-ray Computed Tomography (CT) reconstruction has significantly evolved with the integration of data-driven methods, particularly deep learning. These methods address complex inverse problems that arise from modern imaging hardware advancements, including multi-energy detectors, phase-contrast imaging, and non-standard acquisition geometries. To solve these challenges, it is necessary to develop frameworks that combine classical reconstruction techniques with learning-based approaches.

Several open-source libraries, such as the ASTRA-toolbox \cite{astra}, Core Imaging Library \cite{cil1, cil2}, TIGRE \cite{tigre}, and elsa \cite{elsa}, provide tools for CT reconstruction. However, researchers face difficulties when using these libraries within deep learning workflows. One major challenge is the need for differentiable operators, which are essential for gradient-based optimization. While recent tools like TorchRadon \cite{torchradon} and curadon \cite{curadon} have introduced differentiable projector layers in PyTorch, they still lack comprehensive support for 3D geometries, advanced acquisition trajectories, and artifact simulation.

The emergence of deep learning has transformed tomographic reconstruction. Traditional methods relied on iterative algorithms based on explicit physical models. Although these methods were effective, they struggled with the increasing complexity and data volume of modern imaging setups. In contrast, deep learning provides flexibility and adaptability, enabling end-to-end solutions that combine data-driven insights with physical constraints.

Differentiable operators are essential for integrating physics-based models with machine learning techniques. These operators enable gradient-based optimization within computational graphs, making it possible to train neural networks that incorporate physical modeling in an end-to-end manner. In CT reconstruction, this approach allows the projection and back-projection steps to be embedded directly into the learning pipeline. This enables networks to learn parameters that improve reconstruction quality while adhering to physical principles.

To address these challenges, we introduce an updated version of PYRO-NN \cite{pyronn}, a Python-based library initially designed for TensorFlow. PYRO-NN enables the seamless integration of differentiable CT reconstruction operators into deep learning frameworks. With the initial version of PYRO-NN, several applications have been realized, including reconstruction algorithms for arbitrary CT trajectories \cite{Ye2024DRACODR}, integration of the Defrise algorithm with learnable components \cite{ye2024deep}, enhancement of cephalometric landmark detection in CBCT images \cite{viriyasaranon2024georefinenet}, and applications in sinogram-based defect localization \cite{zhou20242d}. PYRO-NN also contributed to new loss functions and Fourier series learning \cite{sun2024eagle,sun2024data}, achieved dual domain denoising \cite{wagner2023benefit}, and enabled scatter estimation in projection domains \cite{michen2022deep}. These applications showcase PYRO-NN's versatility in data generation, reconstruction, and end-to-end learning. Our updated PYRO-NN framework includes:
\begin{itemize}
\item Native CUDA kernel support for both PyTorch and TensorFlow, enabling efficient projection and back-projection operations for parallel, fan, and cone-beam geometries.
\item Advanced simulation tools for realistic imaging artifacts, such as detector jitter, Poisson noise, Gaussian noise, ring artifacts, and gantry motion blur.
\item Flexible tools for defining and simulating arbitrary acquisition trajectories, allowing modeling of non-standard scanning setups.
\end{itemize}

With these updates, PYRO-NN facilitates the integration of classical and learning-based reconstruction techniques. It supports workflows that aim to reduce redundancy, improve reproducibility, and address the need for more flexible CT reconstruction tools. The addition of PyTorch support, along with new features for trajectory modeling and artifact simulation, enhances the utility of PYRO-NN for researchers and practitioners in CT reconstruction.

\section{The PYRO-NN Framework}  

The PYRO-NN framework provides a flexible and efficient platform for integrating X-ray Computed Tomography (CT) reconstruction operations into modern deep learning workflows. It was originally designed to incorporate native C++ and CUDA-based algorithms into TensorFlow, but recent updates have extended support to PyTorch. This extension enhances usability for the broader machine learning community by enabling seamless integration with one of the most widely used deep learning frameworks.  

PYRO-NN supplies projection and reconstruction operations, including parallel-, fan-, and cone-beam geometries, as native CUDA implementations. These operations are provided as TensorFlow and PyTorch layers, enabling their use within fully end-to-end trainable neural networks. The framework inherently computes analytical gradients for all operators with respect to their inputs, facilitating gradient-based optimization in deep learning pipelines. Additionally, PYRO-NN supports filtered-backprojection (FBP) reconstructions by including filters and weights based on well-established algorithms from scientific literature.  

The high-level Python API provided by PYRO-NN simplifies access to these features. It includes helper functions to automatically manage the computation of gradients and offers flexibility in defining CT reconstruction workflows. Inspired by the CONRAD framework, PYRO-NN also facilitates the integration of data from real clinical scanners. Tools and phantoms available through PyConrad can further extend the functionality of the framework, enabling its application in various research scenarios.

\subsection{Modular Layer Design}  
The core of PYRO-NN’s design lies in its modular implementation of forward- and backprojection operators as native CUDA kernels. These operators are embedded directly into TensorFlow and PyTorch as native custom layers, enabling fine control over device resources such as memory utilization and computation efficiency. Unlike Python-level wrappers, this approach provides precise control over implementation details, ensuring optimal performance.  

The forward-projection operation is implemented using a ray-driven approach, where rays are cast through the volume for each detector pixel, accumulating absorption values along their path. Backprojection, implemented as a voxel-driven operation, reconstructs the data by projecting each voxel onto the detector and accumulating the measured line integrals. These CUDA kernels support 2D parallel- and fan-beam geometries, as well as 3D cone-beam geometries. For 3D cone-beam operations, calibrated projection matrices from real systems can be used.

The framework provides flexibility in memory management by offering both texture-based and kernel-based interpolation methods for 3D cone-beam operations. Texture interpolation ensures faster runtime but requires additional memory, while kernel interpolation is slower but more memory efficient. This adaptability allows users to optimize the framework for their specific computational and memory constraints.  

\begin{figure*}[ht]
    \centering
    \begin{subfigure}{0.16\textwidth}
        \includegraphics[width=\textwidth]{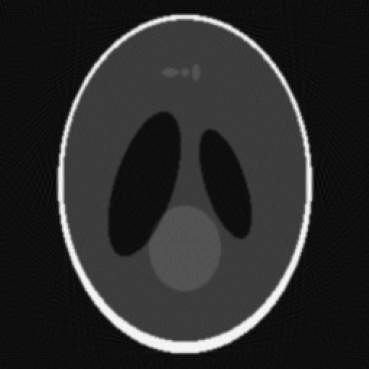}
        \caption{Noise free}
    \end{subfigure}
    \begin{subfigure}{0.16\textwidth}
        \includegraphics[width=\textwidth]{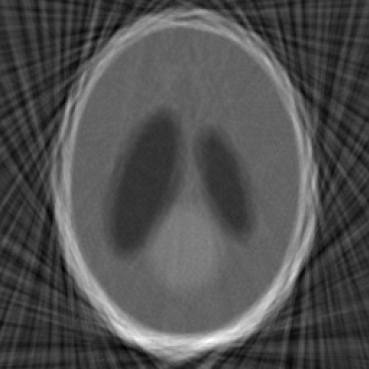}
        \caption{Jitter}
    \end{subfigure}
    \begin{subfigure}{0.16\textwidth}
        \includegraphics[width=\textwidth]{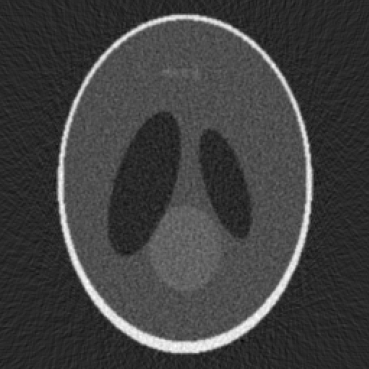}
        \caption{Poisson Noise}
    \end{subfigure}
    \begin{subfigure}{0.16\textwidth}
        \includegraphics[width=\textwidth]{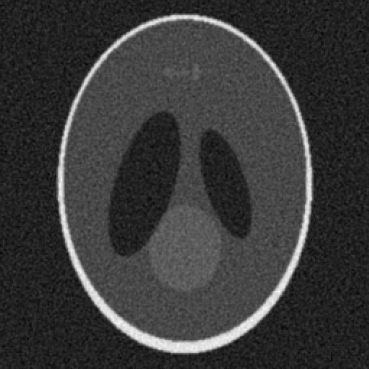}
        \caption{Gaussian Noise}
    \end{subfigure}
    \begin{subfigure}{0.16\textwidth}
        \includegraphics[width=\textwidth]{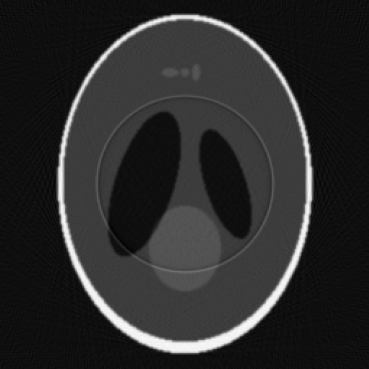}
        \caption{Ring Artifact}
    \end{subfigure}
    \begin{subfigure}{0.16\textwidth}
        \includegraphics[width=\textwidth]{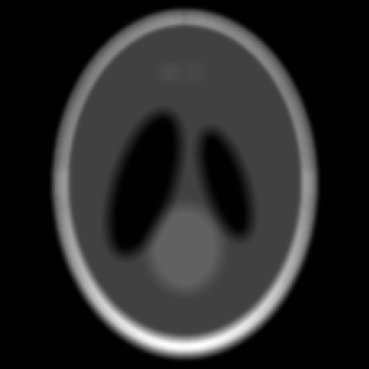}
        \caption{Blur}
    \end{subfigure}
    \caption{Visualization of simulated artifacts with Shepp Logan Phantom.}
    \label{fig:sechs_bilder}
\end{figure*}

\subsection{CT Reconstruction in Neural Networks}  
The PYRO-NN framework enables the embedding of CT acquisition and reconstruction procedures into neural networks using discrete linear algebra. The system geometry is represented as a forward-projection operator \( A \), with its adjoint (backprojection) operator \( A^\top \). Together, these operators allow the implementation of filtered-backprojection (FBP) as:  
\[
x = A^\top F^\top K F p,
\]  
where \( p \) represents the acquired projection data, \( K \) is the diagonal filter matrix applied in the Fourier domain, and \( F \) and \( F^\top \) are the Fourier transform and its inverse, respectively. These operations are fully differentiable, ensuring the flow of gradients throughout the entire network and enabling end-to-end training of neural networks that incorporate physical modeling.  

For computational efficiency, the projection and backprojection operators are not stored as large system matrices but are instead computed on the fly using ray- and voxel-driven algorithms. This approach minimizes memory requirements while maintaining high performance. By supporting these operations in both TensorFlow and PyTorch, the framework now provides flexibility for researchers to work with their preferred deep learning framework. 

\subsection{Code Example: Filtered Backprojection in PYRO-NN}  

The following example demonstrates how PYRO-NN can be used to implement a filtered backprojection (FBP) pipeline for CT reconstruction using the provided layers. The decision between tensorflow or pytorch for computation is done with importing the respective layers:  

\begin{lstlisting}[language=Python, caption={Example of filtered backprojection in PYRO-NN.}, label={lst:pyronn_fbp}, basicstyle=\ttfamily\footnotesize, frame=single, columns=fullflexible, numbers=none]
# ------------ Declare Parameters --------------

# Volume Parameters:
params = dict()
params["volume_shape"] = [256, 256, 256]
params["volume_spacing"] = [0.5, 0.5, 0.5]

# Detector Parameters:
params["detector_shape"] = [400, 600]
params["detector_spacing"] = [1, 1]

# Trajectory Parameters:
params["number_of_projections"] = 360
params["angular_range"] = 2 * np.pi

params["sdd"] = 1200  # Source-Detector Distance
params["sid"] = 750   # Source-Isocenter Distance

# Create Geometry class
geom = GeometryCone3D(**params)
geom.set_trajectory(circular_trajectory_3d(**params)

# Get Phantom 3D
phantom = shepp_logan.shepp_logan_3d(params["volume_shape"])
# Add required batch dimension
phantom = np.expand_dims(phantom, axis=0)  

# ------------------ Call Layers ------------------

# Compute Sinogram
sinogram = ConeProjectionFor3D().forward(phantom, geom)

# Apply Filter for FBP
reco_filter = shepp_logan_3D(**params)
x = fft_and_ifft(sinogram, reco_filter)

# Backprojection
reco = ConeBackProjectionFor3D().forward(x, geom)
\end{lstlisting} 

In this example, the projection and backprojection operators are used to compute the sinogram, apply the Shepp-Logan filter, and reconstruct the volume from the filtered sinogram. The geometry setup ensures that the operators are configured for a 3D cone-beam CT system. This demonstrates the modular and intuitive use of PYRO-NN layers for implementing CT reconstruction workflows.

\subsection{Framework Updates and Enhancements}  
Recent updates to PYRO-NN introduce several key features that significantly enhance its usability, functionality, and versatility. These improvements are designed to meet the evolving needs of researchers and practitioners working in CT reconstruction and related fields.

\paragraph{PyTorch Integration}  
One of the most notable updates is the integration of native PyTorch layers for all projection and reconstruction operations, in addition to the previously supported TensorFlow framework. This expansion makes the framework accessible to a broader audience, as PyTorch has become a widely adopted deep learning library. By supporting both TensorFlow and PyTorch, PYRO-NN ensures flexibility for researchers who may have workflow preferences based on specific projects or existing dependencies.

\paragraph{Advanced Simulation Tool}
The framework now includes advanced tools for simulating realistic imaging artifacts. These include features such as detector jitter, Poisson noise, Gaussian noise, ring artifacts, and gantry motion blur. The ability to simulate such artifacts enables the training and validation of deep learning models under realistic imaging conditions, leading to more robust and generalizable reconstructions. These tools also allow researchers to explore the impact of different types of noise and artifacts on reconstruction quality in a controlled environment. 

\textit{Detector jitter} represents a misalignment in detector readings caused by slight mechanical or electronic fluctuations. In software, random pixel shifts are introduced for each detector column or row to replicate this misalignment. This process can highlight the sensitivity of a reconstruction algorithm to geometric variations.

\textit{Poisson noise} arises from the discrete nature of photon counting. Each pixel value in the sinogram is used as a rate parameter for the Poisson process, which approximates the randomness of detected X-ray photons. This addition of noise is especially important for low-dose simulations, since quantum fluctuations are more noticeable at reduced intensities.

\textit{Gaussian noise} is introduced to model electronic and thermal variations that occur in acquisition hardware. This is achieved by sampling values from a normal distribution with a specified mean and standard deviation, then adding those samples to the sinogram. Gaussian noise is often useful in simulating readout errors and other consistent deviations in detector systems.

\textit{Ring artifacts} result from defective or improperly calibrated detector elements. In the simulation, selected detector columns are set to zero or modified over a segment of the sinogram, which yields ring patterns when the volume is reconstructed. This allows algorithms to be tested on scenarios involving hardware disruption and calibration imbalances.

\textit{Gantry motion blur} comes from gantry rotation during the image acquisition process. To simulate this effect, a rotating kernel is applied to each projection according to its acquisition angle. This convolution-based approach reproduces motion-induced blurring and can be tuned by adjusting the kernel size and shape, reflecting different rotation speeds or mechanical issues.

\paragraph{Flexible Trajectory Modeling}  
PYRO-NN introduces modules for defining and simulating arbitrary acquisition trajectories, expanding its utility to applications involving non-standard scanning geometries. One of the key advancements is the integration of a new method for generating arbitrary projection matrices based on geometric parameters and 3D coordinates of the source and detector. This approach requires only the source and detector positions, as well as the detector directions, to compute the projection matrices.
The implementation allows for both circular and arbitrary trajectories. For circular trajectories, the algorithm initializes the source and detector positions and directions, then computes the rotation incrementally to generate the projection matrices. For arbitrary trajectories, the method directly uses the source and detector positions and orientations as input. The projection matrix is constructed using the source-to-detector vector, scaled and normalized detector directions, and transformations to align the detector with the appropriate coordinate system. 
This method enables a flexible and efficient calculation of projection matrices and allows researchers to model arbitrary trajectories. The approach is based on the work presented by Graetz \cite{graetz2021auto}. By incorporating this functionality, PYRO-NN enables the simulation of complex acquisition setups, making it a valuable tool for developing advanced CT reconstruction techniques and investigating unconventional scanning geometries.
Together, these updates solidify PYRO-NN as a robust and adaptable tool for CT reconstruction research, bridging the gap between classical approaches and modern machine learning techniques. The enhanced functionality, combined with a focus on usability, makes the framework a valuable resource for a wide range of applications.

\section{Conclusion and Future Directions}  

The PYRO-NN framework is an open-source project released under the Apache 2.0 license, ensuring accessibility and reproducibility in modern CT reconstruction research. Its compatibility with both TensorFlow and PyTorch provides flexibility for researchers working with diverse deep learning workflows, while its modular architecture facilitates seamless integration into end-to-end pipelines. The inclusion of native CUDA implementations and analytical gradients ensures efficient and scalable performance across various applications.

Future developments aim to extend PYRO-NN’s functionality, including further algorithmic optimizations, enhanced support for unconventional acquisition geometries, and expanded simulation tools for realistic imaging conditions. These enhancements will enable the framework to address new challenges in signal reconstruction and related domains.

By bridging traditional CT reconstruction techniques with state-of-the-art deep learning methodologies, PYRO-NN offers a practical and versatile solution for advancing research.

\printbibliography
\end{document}